%% file: paper.tex
\documentclass[conference]{IEEEtran}
\IEEEoverridecommandlockouts
\usepackage{cite}
\usepackage{amsmath,amssymb,amsfonts}
\usepackage{algorithmic}
\usepackage{graphicx}
\usepackage{textcomp}
\usepackage{xcolor}
\usepackage{todonotes}

\def\BibTeX{{\rm B\kern-.05em{\sc i\kern-.025em b}\kern-.08em
    T\kern-.1667em\lower.7ex\hbox{E}\kern-.125emX}}
\begin{document}

\title{Ensuring Safe Autonomy: \\ Navigating the Future of Autonomous Vehicles}

\author{\IEEEauthorblockN{Patrick Wolf}
\IEEEauthorblockA{\textit{Department of Safety Engineering}, \textit{Fraunhofer IESE}\\
Kaiserslautern, Germany \\
patrick.wolf@iese.fraunhofer.de}
}

\maketitle

\begin{abstract}
Autonomous driving vehicles provide a vast potential for realizing use cases in the on-road and off-road domains.
Consequently, remarkable solutions exist to autonomous systems' environmental perception and control.
Nevertheless, proof of safety remains an open challenge preventing such machinery from being introduced to markets and deployed in real world.

Traditional approaches for safety assurance of autonomously driving vehicles often lead to underperformance due to conservative safety assumptions that cannot handle the overall complexity.
Besides, the more sophisticated safety systems rely on the vehicle's perception systems.
However, perception is often unreliable due to uncertainties resulting from disturbances or the lack of context incorporation for data interpretation.

Accordingly, this paper illustrates the potential of a modular, self-adaptive autonomy framework with integrated dynamic risk management to overcome the abovementioned drawbacks.
\end{abstract}


\input{introduction}
\input{related_work}
\input{concept}
\input{conclusion}

\bibliographystyle{IEEEtran}
\bibliography{paper}

\end{document}

%% file: introduction.tex
\section{Introduction}
\label{sec:introduction}

Autonomous driving machinery has the potential to transform or disrupt use cases and processes in multiple domains by reducing emissions, saving resources, boosting quality, and increasing safety.
Consequently, industry and research investigate vast resources to develop autonomous machines, leading to exceptional autonomous systems with advanced environmental perception and control capabilities \cite{Okuda14, Yurtsever20}.

Despite the ability to drive and work autonomously, a robot must be proven to be safe before deployment.
However, proving safety and security is an essential and ongoing but open challenge \cite{Mariani18}, \cite{Wang20}, \cite{Schneider24}.
A central aspect is the overall complexity of situations and the limited perceptive reliability.
While safety engineering approaches in the past primarily dealt with components and system functionality and could rely on a human as the final safety guard, this is no longer true for future systems.
Accordingly, the safety of autonomous vehicles depends on external circumstances, which are often unpredictable and not controlled by the system.
Therefore, the analysis of the current situation must regard context information for correct data interpretation and sophisticated decision-making.
Exemplary context data includes location, time, past experiences, or task-related information.

In contrast to state-of-the-art robots, autonomy, and safety strategies, humans are competent in adapting to unforeseen circumstances.
They solve complex perception and control tasks even in unaccounted and surprising situations.
Corresponding examples from literature illustrate that human driving is often considered unsafe from a safety perspective \cite{Naumann21}.
Still, no or only a few accidents occur.
A similar observation was made for perceptive performance where human-inspired modeling significantly outperforms state-of-the-art perception~\cite{Wolf22}, indicating a need for sophisticated safety assessment.

Accordingly, this contribution investigates human-inspired strategies for autonomous vehicle development and concerns the potential of incorporating dynamic risk management into behavior-based systems to overcome the abovementioned challenges.
Sections~\ref{sec:risk_management}--\ref{sec:robotics} discuss relevant methods from safety engineering and behavior-based robotics.
Subsequently, Section~\ref{sec:concept} highlights the potential of combining both approaches, which is concluded by a discussion and outlook (Section~\ref{sec:conclusion}).

%% file: related_work.tex
\section{Risk Management of Autonomous Vehicles}
\label{sec:risk_management}

Runtime safety assurance of autonomous vehicles has different facets, such as nominal and functional safety.
The latter is covered well by safety standards \cite{Tiusanen20}.
The authors of \cite{ShalevShwartz18} highlight that functional safety is necessary but insufficient to ensure overall safety.
Consequently, nominal safety is crucial for ensuring safety but needs to be sufficiently standardized.
Accordingly, introducing the Responsibility-Sensitive Safety (RSS) concept \cite{IEEE2846} provided a set of common sense rules for safety in the on-road domain that is widely adopted.
The off-road domain has comparable approaches, such as those presented by \cite{Ropertz18a}.

The authors of \cite{Reich20} recognized the relevance of situational awareness for safe decision-making and proposed a situation-aware dynamic risk management concept.
Worst-case safety assumptions lead to a systematic underperformance of autonomous agents.
Therefore, they introduce a sophisticated, situation-specific safety concept with tailored control parameter adaptation for more granular decision-making.

Still, the integration into nominal driving frameworks remains an open challenge, including the dependency on nominal perception.
Even though decision-making can consider the broader circumstances, these decisions rely on an uncertain perception.

\section{Behavior-based Robotics}
\label{sec:robotics}

A proper method for creating safe and adaptive autonomous systems is the behavior-based paradigm.
Behavior-based systems are highly robust, fault-tolerant, and can realize robotic control and perception systems.
Fundamental properties are modularization, behavior interaction, parallelism, multi-goal-following, and redundancy.
Unlike sense-plan-act architectures, behavior networks decompose functionality into different behaviors, and decision-making occurs decentralized on a component base.


Behavior trees \cite{Colledanchise18} are widespread for control realization but cannot address perceptive requirements.
Besides, behavior networks, such as the iB2C framework \cite{Wolf22}, excel in both domains by following the principle of task decomposition and multi-goal following.
Behavior networks separate data flow from arbitration flow.
This approach, including continuous network arbitration, allows for realizing non-discrete states, a key benefit.
Consequently, behavior networks are exceptional at trading off contradicting goals and properties.

Behavior networks are a powerful and established method for handling robot control and perception.
Nonetheless, the approach does not cover systematic risk assessment and industrial standardization.

%% file: concept.tex
\section{Towards Dependable Behavior}
\label{sec:concept}

Robotic and autonomous systems are commonly understood as sense-plan-act systems in the automotive domain:
Sensory stimuli lead to environmental representations used for planning and plan execution.
A significant shortcoming with this understanding is the lack of dynamic knowledge incorporation into this process.
Accordingly, it is easy to identify situations where (reactive) nominal safety fails, as the following two examples depict:
\begin{LaTeXdescription}
\item[Scenario 1] An autonomous on-road vehicle drives on a highway and uses RSS as the nominal safety function. 
The vehicle follows another vehicle at a safe distance.
Now, the front vehicle approaches a broken-down truck standing in the same lane.
The front vehicle does not brake and changes lanes immediately before a crash happens.
Abruptly, the nominal safety of the ego car is confronted with a non-safe braking distance to the still-standing truck.
Seemingly, a safe reaction would consider the broader circumstances to proactively act safely and consider the truck.
\item[Scenario 2] An autonomous off-road vehicle climbs an inclination when a previously non-visible terrain feature is identified as an obstacle.
Thus, collision avoidance intervenes in navigation and slows down to prevent the collision.
As a result, the vehicle loses traction and starts sliding, with the danger of rolling over.
The safe reaction in this situation strongly depends on the precise circumstances, and a trade-off between different risks is necessary.
Therefore, it can be more reasonable to crash and bring the vehicle to a safe state later when the terrain allows it, which is a strategy applied by expert drivers.
\end{LaTeXdescription}

Apparently, a counterargument to the above-described safety shortcomings could be that the specification of these examples was incomplete and, accordingly, safety was poorly designed.
Nonetheless, adding extra cases and specific situations to safety strategies alone is insufficient to create safety machines.
Specification will always be incomplete since changes happen, new circumstances arise, or technology evolves.
Accordingly, a restricted or reduced view of a problem statement is inadequate from a system perspective, which must be able to handle new situations.
Solving safety issues from a component view, such as using a radar sensor for distance determination for collision avoidance, ignores the big picture and can only function in minimal situations. 

%% file: conclusion.tex
\section{Conclusion}
\label{sec:conclusion}

In summary, this paper highlighted the challenges of runtime safety assurance for autonomous vehicles including the relevance of situational awareness for risk assessment, perceptive performance, and decision-making.

This contribution claims that dynamic risk management can significantly benefit from incorporating methodologies of behavior-based systems.
Initial attempts at situation-based risk assessment in behavior-based autonomy already showed promising results \cite{Hamza22}.

Future work addresses the systematic incorporation of the approaches, including technological demonstration and standardization.